\pdfoutput=1

\documentclass[11pt]{article}

\usepackage[final]{acl}

\usepackage{times}
\usepackage{latexsym}
\usepackage{amssymb}
\usepackage{bbm}
\usepackage{array}
\usepackage{hyperref}
\usepackage{tablefootnote}
\usepackage{tikz}
\usepackage[edges]{forest}
\usetikzlibrary{mindmap, shadows}
\usetikzlibrary{arrows,shapes,positioning,shadows,trees,arrows.meta}

\tikzset{
  basic/.style  = {draw, rounded corners=2pt, font=\small, thin, align=center, text width=8cm, drop shadow, rectangle}
}

\usepackage[T1]{fontenc}

\usepackage[utf8]{inputenc}

\usepackage{microtype}

\usepackage{inconsolata}

\usepackage{graphicx}

\title{A Survey on Feedback-based Multi-step Reasoning for Large Language Models on Mathematics}

\author{Ting-Ruen Wei \\
  Santa Clara University \\
  \texttt{twei2@scu.edu} \\\And
  Haowei Liu \\
  Santa Clara University \\\AND
  Xuyang Wu \\
  Santa Clara University \\\And
  Yi Fang \\
  Santa Clara University \\
  \texttt{yfang@scu.edu} \\
}

\begin{document}
\maketitle
\begin{abstract}
Recent progress in large language models (LLM) found chain-of-thought prompting strategies to improve the reasoning ability of LLMs by encouraging problem solving through multiple steps. Therefore, subsequent research aimed to integrate the multi-step reasoning process into the LLM itself through process rewards as feedback and achieved improvements over prompting strategies. Due to the cost of step-level annotation, some turn to outcome rewards as feedback. Aside from these training-based approaches, training-free techniques leverage frozen LLMs or external tools for feedback at each step to enhance the reasoning process. With the abundance of work in mathematics due to its logical nature, we present a survey of strategies utilizing feedback at the step and outcome levels to enhance multi-step math reasoning for LLMs. As multi-step reasoning emerges a crucial component in scaling LLMs, we hope to establish its foundation for easier understanding and empower further research.
\end{abstract}

\section{Introduction}
Large language models (LLMs) have made significant progress and demonstrated improvements on many benchmarks \citep{hendrycks2021measuring}. Traditional methods to improve the model performance include increasing the size of the dataset and enhancing its quality to approximate the population distribution. However, humans usually do not require as many practice problems as a training set to be able to solve the same type of problems. Based on this motivation, researchers have turned to enhancing LLMs through multi-step reasoning. Instead of learning from a vast amount of training data through optimizing a training objective, reasoning decomposes a complex problem into multiple simpler problems and makes decisions based on the context and their patterns. On the other hand, chain-of-thought (CoT) 
\begin{figure}[!h]
  \includegraphics[width=\columnwidth]{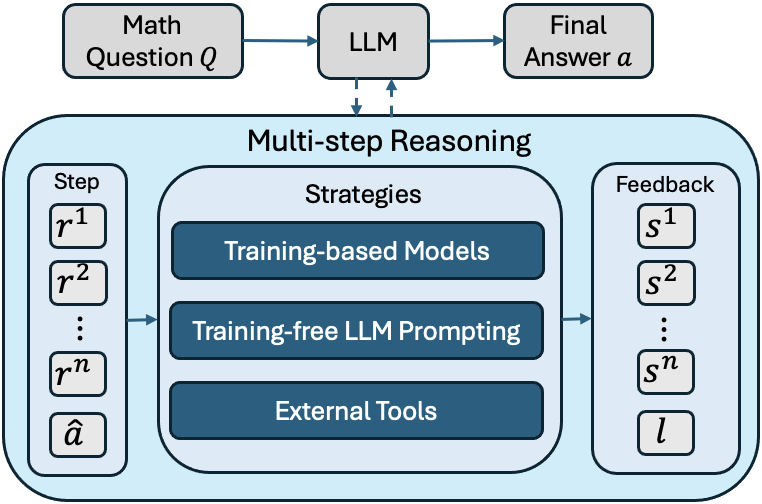}
  \caption{Our survey coverage on feedback-based multi-step reasoning. We focus on strategies that utilize feedback for enhancing the LLM performance across multiple steps on math problems.}
  \label{fig:overview}
\end{figure}
\citep{wei2022chain} as a pioneering work, has found great success through prompting LLMs to output step-by-step responses, establishing the paradigm of multi-step reasoning \citep{chu2024navigate} and opening new avenues for training LLMs to produce such structured outputs, which showed better performance than prompting alone \citep{uesato2022solving}. This is empowered by step-level feedback (process rewards) and outcome-level feedback (outcome rewards) which can be seen as a special case of step-level feedback with no intermediate rewards. Step-level feedback encourages correct reasoning steps and sets the solution on the right track, and such feedback can be provided by training-based models, training-free LLM prompting, or external tools, as shown in Figure \ref{fig:overview}. Due to the logical nature of mathematics, lots of research apply reasoning on LLMs to solve math problems, creating the need for a survey. The specific focus on multi-step reasoning is due to the trend of solving problems by guiding LLMs through multiple steps with feedback, which is different from traditional methods (CoT and supervised fine-tuning (SFT)) where there is no step-level feedback. In this paper, we present a survey on not only fine-tuning approaches but also training-free strategies that guide LLMs through multiple steps to solve math problems.

Previous survey papers for LLMs in reasoning tend to stay on the general domain. While \citet{huang2023towards} discussed an overview of LLM reasoning and \citet{qiao2023reasoning} focused on prompting strategies for general reasoning, \citet{guan2024search} proposed verifier engineering. On math, \citet{liu2023mathematical} addressed math language models and \citet{ahn2024large} discussed the datasets and general methods such as prompting frozen LLMs and fine-tuning. Close to our work, \citet{plaat2024reasoning} covered multi-step math reasoning but with a focus on prompt-based strategies, overlooking a mainstream of fine-tuning strategies utilizing feedback that can lead to better performance \citep{uesato2022solving}. In this survey, we present training-free and training-based strategies utilizing feedback to guide LLMs in multi-step reasoning on math.


\begin{figure}[!t]
  \includegraphics[width=\columnwidth]{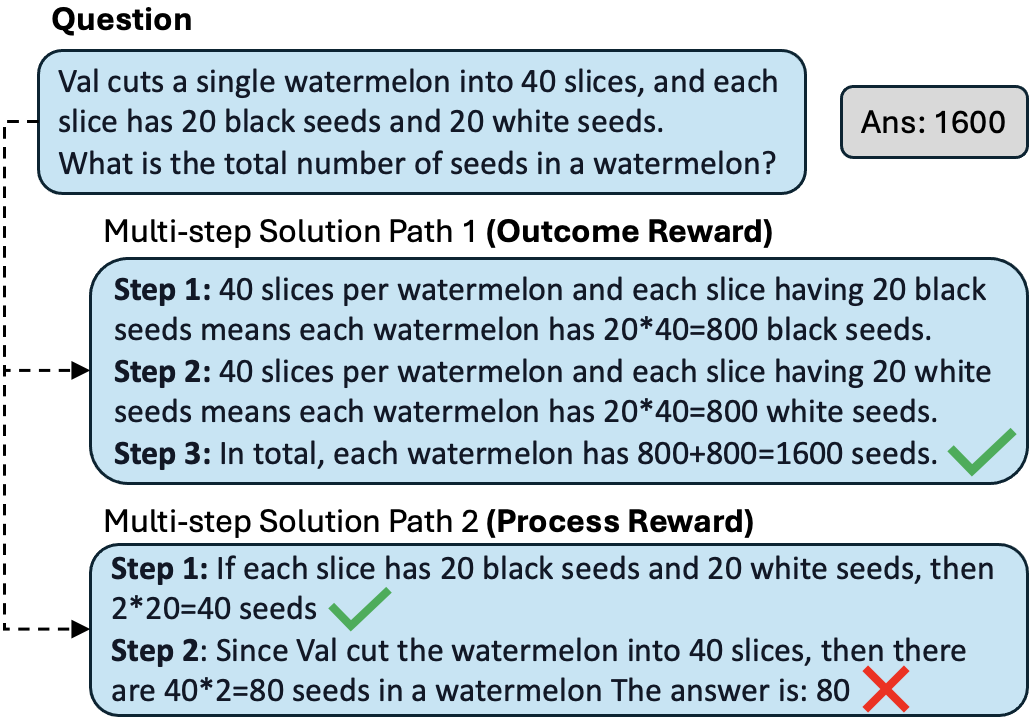}
  \caption{Illustrative example. Outcome reward evaluates the outcome (GSM8K dataset \citep{cobbe2021training}), while process reward evaluates each reasoning step (Math-Shepherd dataset \citep{wang2024math}).}
  \label{fig:example}
\end{figure}

\section{Background}

\subsection{Problem Statement}
Given a question $Q$, an LLM generates the $m^{th}$ solution consisting of many reasoning steps denoted as $r^1_m, \dots, r^n_m$ and a final answer $\hat{a}_m$, resulting in a set of solution paths $\{r^1_m, \dots, r^n_m, \hat{a}_m\}_{m=1}^M$ where $n$ and $M$ are the total number of steps and solutions. Each final answer $\hat{a}_m$ has a correctness label $l_m$, and the final prediction $a$ is selected from the set of final answer(s) $\{\hat{a}_m\}_{m=1}^M$ and compared against the correct final answer $A$ for correctness which is the objective to maximize.

\subsection{Preliminaries}
To integrate multi-step reasoning into LLMs, \citet{uesato2022solving} conducted SFT on a pre-trained LLM by treating the question as input tokens and the multi-step solution as the target tokens. They annotated each step whether the $m^{th}$ solution has been correct so far, resulting in step-label pairs $\{(r^i_m, s^i_m)\}_{i=1}^n$, where $s$ is a binary label. Process reward models (PRMs) utilize this additional supervision for each reasoning step and the training combines with the language model objective. In \citet{cobbe2021training}, a scalar head implemented as a bias parameter and a gain parameter operates on the logit of the special token for binary classification. To save annotation cost, all reasoning steps can instead be labeled as whether the final answer is correct, resulting in outcome reward models (ORMs). Figure \ref{fig:example} illustrates the difference between process and outcome rewards with an example. Though \citet{uesato2022solving} found comparable performance between PRMs and ORMs, \citet{lightman2023let} conducted a similar experiment and found PRMs to outperform ORMs by a large margin, attributing the success to the larger size of the dataset. 

\begin{figure*}[t]
\begin{forest}
    for tree={
      draw, semithick, rounded corners, drop shadow,
     fill=blue!20,
     font=\scriptsize,
       text width = 19mm, text badly centered,
             edge = {draw, semithick},
           anchor = east,
             grow = east,
    forked edge,            
            s sep = 4mm,    
            l sep = 4mm,    
         fork sep = 4mm,    
               }
[Strategies, rotate=90
    [Training-free
        [By External Tool
            [MathDivide \citep{srivastava2024mathdivide}; DTV \citep{zhou2024don}, font=\scriptsize, text width=95mm, fill=white]]
        [By LLM Logits
            [\citet{vacareanu2024general}; PathFinder \citep{golovneva2023pathfinder}; LeCo \citep{yao2024learning}; RAP \citep{hao2023reasoning}; TreeBoN \citep{qiu2024treebon}; UAG \citep{yin2024reasoning}, font=\scriptsize, text width=95mm, fill=white]]
        [By LLM Response
            [\citet{xie2024self}; Natural Program \citep{ling2024deductive}; LoT \citep{zhao2024enhancing}; CoT Rerailer \citep{wan2024cot}; SelfCheck \citep{miao2023selfcheck}; SSC-CoT \citep{zhao2024stepwiseselfconsistentmathematicalreasoning}, font=\scriptsize, text width=95mm, fill=white]]
    ]
    [Training-based, text width=19mm
        [Step- and Outcome-level Feedback[
        CoRe \citep{zhu2023solving}; \citet{setlur2024rewarding}, font=\scriptsize, text width=95mm, fill=white
        ]]
        [Outcome-level Feedback
            [Rule-based[
            DeepSeekMath \citep{shao2024deepseekmath}; DeepSeek-R1 \citep{guo2025deepseek}, font=\scriptsize, text width=70mm, fill=white]
            ]
            [Generative[
            GenRM \citep{zhang2024generative}; STILL-1 \citep{jiang2024technical}, font=\scriptsize, text width=70mm, fill=white]
            ]
            [Discriminative[
            \citet{cobbe2021training}; TinyGSM \citep{liu2023tinygsm}; V-STaR \citep{hosseini2024v}; REPS \citep{kawabata2024rationale}; OVM \citep{yu2024ovm}; \citet{brown2024large}; TS-LLM \citep{feng2023alphazero}; GraphReason \citep{cao2023enhancing}, font=\scriptsize, text width=70mm, fill=white]
            ]
        ]
        [Step-level Feedback
            [Refinement[
            LeMa \citep{an2023learning}; \citet{jiao2024learning}; Step-DPO \citep{lai2024stepdpo}; DAPO \citep{liu2024improving}; StepCo \citep{wu2024enhancing} LM2 \citep{juneja2024lm2}; SORM \citep{havrilla2024glore}; REFINER \citep{paul2024refiner}; ReST-MCTS* \citep{zhang2024rest}; \citet{gao2024designing}; BackMATH-LLM \citep{zhang2025backmath}, font=\scriptsize, text width=70mm, fill=white]
            ]
            [Search[
            DBS \citep{zhu2024deductive}; Mindstar \citep{kang2024mindstar}; \citet{xu2023no}; HGS-PRM \citep{ma2023let}; TVM \citep{lee2024token}; Grace \citep{khalifa2023grace}; OREO \citep{wang2024offlinereinforcementlearningllm}; RSD \citep{liao2025RewardGuided}; LE-MCTS \citep{DBLP:journals/corr/abs-2412-15797}; AlphaMath \citep{DBLP:journals/corr/abs-2405-03553}; \citet{snell2024scaling}, font=\scriptsize, text width=70mm, fill=white]
            ]
            [Aggregation[
            DiVeRSe \citep{li2022making}; Math-Shepherd \citep{wang2024math}; ReasonEval \citep{xia2024evaluating}; Self-Explore \citep{hwang2024self}; VerifierQ \citep{qi2024verifierq}; AutoPSV \citep{lu2024autopsv}; MiPS \citep{wang2024multi}; MATH-Minos \citep{gao2024reason}; Tree-PLV \citep{he2024advancing}; OmegaPRM \citep{luo2024improve}; \citet{yuan2024free}; \citet{zhang2025lessons}; \citet{ma2024step}; PQM \citep{DBLP:journals/corr/abs-2410-11287}; ER-PRM \citep{DBLP:journals/corr/abs-2412-11006}, font=\scriptsize, text width=70mm, fill=white]
            ]
        ]
    ]
]
    \end{forest}

\caption{Taxonomy of multi-step math reasoning through feedback. Training-based approaches leverage step- or outcome-level feedback. Step-level feedback can be aggregated for voting or best-of-N, or assist search or refinement. Outcome-level feedback can train discriminative or generative ORMs, or be determined by rules. Training-free techniques evaluate each step by direct LLM responses, LLM token logits, or an external tool to obtain step-level feedback.}
\label{fig:taxonomy}
\end{figure*}
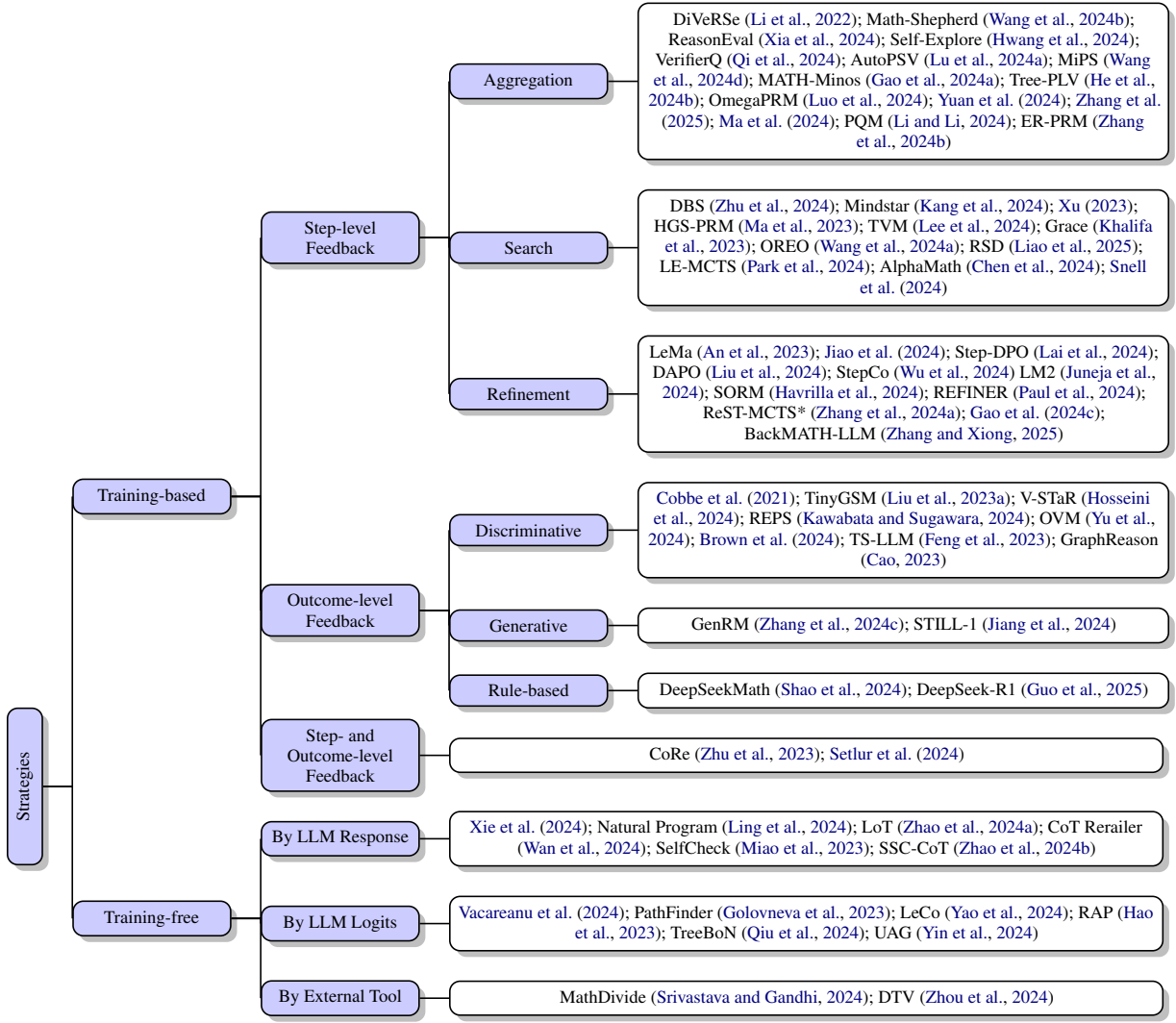

While the LLM generation process typically follows greedy decoding to output tokens with the highest probability, Self-consistency \citep{wang2022self} found success by sampling 40 solutions $\{\hat{a}_m\}_{m=1}^{40}$ and selecting the most consistent answer as the final answer. Consistency can be measured through majority voting or weighted voting. Majority voting selects the most common answer: 

\begin{equation}
a = \arg \max_a \sum_{i=1}^{m} \mathbf{1}(\hat{a}_i = a)
\end{equation}

\noindent while weighted voting selects the final answer with the largest sum of weights: 

\begin{equation}
a = \arg \max_a \sum_{i=1}^{m} w_i \cdot \mathbf{1}(\hat{a}_i = a)
\end{equation}

\noindent where $w_i$ is the weight for $\hat{a}_i$. Alternatively, best-of-N \citep{cobbe2021training, lightman2023let} selects the solution with the highest weight: 

\begin{equation}
a = \hat{a}_{\arg \max_{1 \leq i \leq m} w_i}
\end{equation}

\noindent These strategies can directly follow a trained ORM by setting $w_i$ as the model's scalar output in probability. For PRMs, the model predicts a probability score for each reasoning step, and these scores are aggregated by either taking their product, their minimum, or the last value, to represent $w_i$. In addition, stepwise scores can assist beam search in finding the best solution path \citep{yu2024ovm}. By treating each reasoning step as a node and its process reward as the weight, beam search can sample more flexibly to reach the final answer. 

As there can be multiple paths to solve a question, Monte Carlo Tree Search (MCTS) \citep{browne2012survey} presents a suitable strategy for exploiting and exploring reasoning steps to obtain the best solution path. The tree regards the question input as the root node, each step as a non-leaf node, and each final answer as a leaf node. MCTS selects the node with the hightest value to expand and simulate, and the final result updates the values of the nodes in the path taken. The selection is based on the following heuristic:

\begin{equation}
    v_j + C\sqrt{\ln(N)/N_j}
\end{equation}

\noindent where $v_j$ is the value for node $j$, $C$ is a constant balancing between exploitation and exploration, and $N$ and $N_j$ are the times visited for the current node and for node $j$. From the root node, progressively selecting the child node with the highest value results in the final solution path.

While some strategies involve training a reward model, there are also training-free approaches that do not train a model to evaluate each reasoning step. Based on the way of evaluation, they are categorized into using the LLM's direct response, the LLM's logits, and an external tool. We present the taxonomy in Figure \ref{fig:taxonomy}.

\section{Training-based Strategies}

\subsection{Step-level Feedback}
On top of SFT, some models enhance the reasoning process by training on stepwise labels to evaluate the quality of a reasoning step. Such trained models utilize the label predictions as scores to guide or select the best solution path. Specifically, some aggregate stepwise scores to represent the entire solution in a voting scheme, while some others utilize stepwise scores to guide tree search, and others refine or prompt refinement from LLMs.

\subsubsection{Aggregation}
It is common to jointly train the language model with a binary cross-entropy term on the correctness of intermediate steps. The aggregation of stepwise scores often follow majority voting, weighted voting, or best-of-N to select the final solution. The key distinction between the papers lies in their ways of defining, annotating, or training with stepwise labels. Based on the motivation that the reasoning steps are not equally wrong, DiVeRSe \citep{li2022making} labels steps that do not occur in paths that lead to the correct final answer as incorrect. Math-Shepherd \citep{wang2024math} defines stepwise labels as the potential to arrive at the correct final answer and automates the labeling process through soft and hard estimation. For each step, an LLM generates multiple solutions, and soft estimation computes the percentage of paths reaching the correct final answer (also in MiPS \citep{wang2024multi}), while hard estimation determines whether the correct answer is achieved at all. Similarly in Self-Explore \citep{hwang2024self}, for a solution path leading to an incorrect final answer, multiple solution paths are generated from each step. The first step that does not have one of its newly generated paths reaching the correct answer is labeled as the first error, forming stepwise labels. ER-PRM \citep{DBLP:journals/corr/abs-2412-11006} proposes entropy-regularized PRMs to balance reward optimization with policy stability, preventing drastic changes from the original policy. In contrast to classification-based PRMs that assess each step independently, PQM \citep{DBLP:journals/corr/abs-2410-11287} captures dependencies among reasoning steps by optimizing their ranking using Plackett-Luce loss \citep{plackett1975analysis, luce1959individual}. This enables more precise stepwise reward distribution and enhances verification accuracy with best-of-N. On the other hand, ReasonEval \citep{xia2024evaluating} presents a metric to score solution paths through their validity and redundancy. AutoPSV \citep{lu2024autopsv} utilizes a trained ORM to label each reasoning step as its probability of leading to the correct final answer. By comparing the change in confidence between steps against a threshold, the steps are labeled as valid or erroneous.

With each node representing a reasoning step and an edge connecting two consecutive steps, some papers structure solution paths as a tree for better exploration. OmegaPRM \citep{luo2024improve} uses MCTS to annotate each reasoning step. The first error in the solution path rollout is determined through binary search. \citet{zhang2025lessons} finds the quality of step annotations done by MCTS to be suboptimal and improves it by incorporating LLM-as-a-judge \citep{li2024llms} for evaluation. \citet{ma2024step} further investigates MCTS for automatic annotation and finds that the thought process in natural language is not important compared to the actual calculation. Tree-PLV \citep{he2024advancing} labels each step through soft estimation and expands on those with the highest values. Upon reaching a correct final answer, the model trains on these reasoning steps with their values through preference instead of scalar values.

Before training on binary stepwise labels, MATH-Minos \citep{gao2024reason} fine-tunes the LLM on stepwise natural language feedback that explains how each step is correct or incorrect. Instead of SFT, VerifierQ \citep{qi2024verifierq} trains the PRM through Q-learning. To avoid step-level annotation costs, \citet{yuan2024free} trains an implicit PRM with labels at the outcome level by parameterizing the reward as a ratio of log-likelihoods between the policy and the reference models.


\subsubsection{Search}
Instead of generating multiple solution paths and selecting the final one through aggregation and voting, stepwise scores can also directly guide the generation of one solution path through search. DBS \citep{zhu2024deductive} implements beam search on reasoning steps using stepwise scores based on their logical consistency with the previous steps. The PRM is specifically trained on margin ranking loss \citep{shashua2002ranking} with synthetic data that simulate typical errors. TVM \citep{lee2024token} trains a PRM with token-level values instead of step-level values to guide beam search during inference. Multiple solution paths are generated, and at each token level, the token value refers to the proportion of paths that a particular token arrives at the correct final answer. On top of beam search, Mindstar \citep{kang2024mindstar} uses the stepwise scores as a heuristic in levin tree search \citep{orseau2018single} which additionally considers the depth of the trajectory. On the other hand, HGS-PRM \citep{ma2023let} applies the greedy search algorithm to select the best solution path with a PRM. The reward model classifies each step as positive, neutral, or negative, descending in priority for greedy search, while negative steps trigger backtracking. \citet{xu2023no} turns LLMs into a residual-based energy model and applies noise contrastive estimation to estimate the energy function’s parameters for guiding MCTS. AlphaMath \citep{DBLP:journals/corr/abs-2405-03553} eliminates the need of human- or GPT-annotated labels and relies on solution paths generated through MCTS. After jointly training the policy and PRM, beam search utilizes the trained PRM to reduce the cost of MCTS at inference. Similarly, OREO \citep{wang2024offlinereinforcementlearningllm} improves upon DPO by jointly training the policy model and a value function to optimize the soft Bellman Equation. The trained value function facilitates beam search at inference. In a comprehensive analysis, \citet{snell2024scaling} finds that under a low computational budget, beam search outperforms best-of-N, but the trend reverses as the budget increases, and look-ahead search consistently underperforms these methods. On the other hand, RSD \citep{liao2025RewardGuided} combines a PRM with speculative decoding to improve the decoding efficiency. Grace \citep{khalifa2023grace} utilizes a discriminator model to score all sampled steps and select the best step to add to the solution path. To train the model, sampled solution paths with incorrect final answers are compared to the reference solution paths to construct contrative pairs. The model trains on max-margin loss \citep{rosasco2004loss} to assign high scores to correct steps and low scores to incorrect steps. Unlike prior methods that rely on a single model, LE-MCTS \citep{DBLP:journals/corr/abs-2412-15797} leverages different LLMs within MCTS and selects the most promising steps dynamically across multiple models, guided by a PRM.

\subsubsection{Refinement}
Besides serving as the score for aggregation or the heuristic for search, stepwise feedback can also directly train LLMs or prompt them for refinement. \citet{jiao2024learning} annotates reasoning steps with a trained PRM and trains the language model on these labels through DPO \citep{rafailov2024direct}. Step-DPO \citep{lai2024stepdpo} extends DPO by optimizing intermediate reasoning steps alongside final outputs, improving multi-step reasoning. Similarly, DAPO \citep{liu2024improving} applies preference learning at the step level instead of the outcome level, encouraging the language model to generate high-quality steps. BackMATH-LLM \citep{zhang2025backmath} utilizes a PRM and another PRM for backward reasoning to refine the model through PPO \citep{schulman2017proximal}. \citet{gao2024designing} stabilizes reinforcement learning by setting an upper bound on the reward to reduce incorrect steps and subtracting the rewards between neighboring steps to discourage repeating similar steps. On the other hand, ReST-MCTS* \citep{zhang2024rest} integrates a PRM with MCTS to collect reasoning steps and their labels to enhance self-training of the language model as the policy model. Positive steps are used to train the language model, while all labels are used to train the PRM. LeMa \citep{an2023learning} fine-tunes LLMs on correction data obtained by prompting the GPT-4 model to correct false solution paths, with difficult questions sampled to expand the original set of questions. SORM \citep{havrilla2024glore} designs stepwise outcome rewards as the probability of the reaching the correct final answer given an intermediate reasoning step. The pipeline incorporates additional models for global and local refinement, drafting an entirely different solution and modifying local parts of the solution.

The stepwise refinement can also be directly prompting an LLM. StepCo \citep{wu2024enhancing} generates one solution path and utilizes a trained PRM to rate each reasoning step. Steps lower than a threshold are considered erroneous and revised by the LLM until all steps exceed the threshold. Similarly, LM2 \citep{juneja2024lm2} decomposes a problem into sub-questions and sequentially verify the answer to each sub-question, and an incorrect sub-answer prompts the re-generation of the sub-question. The verifier model is trained to identify nine types of incorrectness, facilitating stepwise refinement. On the other hand, REFINER \citep{paul2024refiner} trains a generator model to generate solution paths and a critic model to provide feedback to one of the generated paths in an iterative fashion.



\subsection{Outcome-level Feedback}
Instead of step-level feedback, some papers turn to utilize outcome-level feedback, potentially avoiding a higher cost at the expense of more fine-grained feedback. Outcome-level feedback can be obtained by discriminative or generative ORMs or determined by rules.

\subsubsection{Discriminative ORM}
As the pioneering work, \citet{cobbe2021training} trains an ORM to select the best generated solution. A solution generator is first trained to output solutions given a question, and multiple solutions are generated, along with their binary correctness labels, to train the ORM to output the correctness probability of a given solution. For inference, a large number of diverse solutions are generated, evaluated by the trained ORM, and ranked to select the best solution. Many papers follow the same framework with various enhancements. REPS \citep{kawabata2024rationale} treats only the best solution path as positive with all others as negative in training the ORM. Pairs of solution paths are iteratively compared to yield the best one for each problem. V-STaR \citep{hosseini2024v} utilizes the correct and incorrect solutions during each training iteration in STaR \citep{zelikman2022star} to simultaneously train the ORM, taking advantage of the solution generator that is improving throughout the iterations, instead of a fixed generator. Similarly, TS-LLM \citep{feng2023alphazero} proposes a framework to iteratively guide inference with tree search and train the policy model and ORM with the collected trajectories. On the other hand, OVM \citep{yu2024ovm} utilizes a trained ORM to guide beam search in inference by evaluating at the step level and leveraging outcome rewards as process rewards. GraphReason \citep{cao2023enhancing} utilizes an additional graph to aggregate the solution paths with the same final answer to train an answer classifier. At inference, multiple solution paths are sampled and processed in the same way, and the outcome with the highest predicted level of correctness is selected as the final answer. Moreover, TinyGSM \citep{liu2023tinygsm} finds an ORM to improve the performance of a small language model. With a comprehensive analysis, \citet{brown2024large} observes that increasing the number of samples generated boosts the chance of generating the correct solution. However, the chance of identifying that correct solution remains relatively unchanged despite the increase in samples.

\subsubsection{Generative ORM}
ORMs can also be trained through a generative objective, as GenRM \cite{zhang2024generative} trains the verifier model by predicting the next token. Given the problem and a potential solution, the model answers the question: ``\textit{Is the answer correct (Yes/No)?}'' The token probability of ``Yes'' and ``No'' are extracted for ranking at test time. To enhance with CoT, GenRM-CoT adds the verification steps before answering ``Yes'' or ``No''. At test time, multiple verification rationales are generated, and majority voting is applied to their token probabilities of ``Yes''. To gather training data, instead of prompting the LLM to verify step-by-step, reference-guided grading \citep{zheng2023judging} guides the step-by-step verification by providing a reference solution that arrives at the correct final answer. Similarly, STILL-1 \citep{jiang2024technical} trains a generative reward model and utilizes it to evaluate rollouts in MCTS at test time to obtain step-level values which are combined with Self-consistency \citep{wang2022self} estimates.

\subsubsection{Rule-based Rewards}
Instead of training a neural network for an ORM, DeepSeekMath \citep{shao2024deepseekmath} and DeepSeek-R1 \citep{guo2025deepseek} use rules to obtain feedback at the outcome level and GRPO \citep{shao2024deepseekmath} to improve the language model with reinforcement learning. A group of solutions are generated and evaluated for correctness to obtain feedback which is further normalized within the group. This reduces the computational resources during training compared to PPO \citep{schulman2017proximal} and neural ORMs.

\subsection{Step- and Outcome-level Feedback}
Instead of using only step- or outcome-level feedback, some utilize both to enhance reasoning. CoRe \cite{zhu2023solving} trains a PRM and an ORM to label reasoning paths in MCTS and adds high-quality paths back to the dataset for training the reward models. The iterative process assist MCTS to better evaluate each step and obtain the final answer. On the other hand, \citet{setlur2024rewarding} measures the advantage of each step defined as the gain in likelihood of reaching the correct answer by taking that step, and combines with an ORM. Termed process advantage verifiers, the model achieves better reasoning performance and higher efficiency than PRMs and ORMs.

\section{Training-free Approaches}
Instead of training PRMs, ORMs, or fine-tuning the policy model, some papers utilize training-free approaches through frozen LLMs or external tools to obtain feedback in navigating multi-step reasoning.

\subsection{Evaluate by LLM Response}
At each step, the solution path can be evaluated by directing prompting the LLM. Self-evaluation \citep{kadavath2022language} turns an initial response of the LLM into a true-or-false question and prompts the same LLM for confirmation, and \citet{xie2024self} utilizes this result to guide stochastic beam search to find the final solution path. Natural Program \citep{ling2024deductive} verifies each reasoning step against the premises from the question. Multiple solution paths are sampled for the verification process and only those that have all the steps verified are applied majority voting to obtain the final answer. LoT \citep{zhao2024enhancing} generates a solution path and verifies each step for revision. The verification prompts two explanations to why the step is correct and why it is incorrect, and the step is revised if the latter is preferred over the former. Similarly, SelfCheck \citep{miao2023selfcheck} prompts the LLM to verify its own steps by re-generating each step and comparing against it. The results serve as the weight of the solution path in majority voting. On the other hand, CoT Rerailer \citep{wan2024cot} generates multiple solution paths and selects the least hallucinated one for an LLM agent to verify each step. A debate is initiated between multiple LLMs, and a new solution path can be re-generated. SSC-CoT \citep{zhao2024stepwiseselfconsistentmathematicalreasoning} also generates multiple solutions and ensures their logical consistency by finding the overlapping steps with an LLM and querying a knowledge graph, increasing the likelihood of accurate and coherent solutions.

\subsection{Evaluate by LLM Logits}
Besides using the LLM's response, the token logits output by LLMs provide another way to evaluate each reasoning step. RAP \citep{hao2023reasoning} utilizes the token probability of ``Yes'' in self-evaluation along with its confidence as the heuristic to find the solution through MCTS. UAG \citep{yin2024reasoning} detects uncertainty spikes in the solution path at the token level using LLM logits and adds relevant exemplars to the beginning of the prompt to leverage their insights. LeCo \citep{yao2024learning} designs three statistics based on token probabilities to evaluate the confidence of each step. The approach generates an initial solution and accepts all the steps until the lowest-scoring one. The LLM is prompted iteratively with the updated list of steps until two consecutive answers are consistent. Similarly, \citet{vacareanu2024general} integrates perplexity with relevance, accuracy, and logical consistency for diverse evaluations. Each aspect is aggregated by the geometric mean across all steps, and their average represents the weight for a solution path. Using common n-grams to signify consistency, PathFinder \citep{golovneva2023pathfinder} constructs a tree of solution paths. Each step has a corresponding weight for sampling, represented by the summation of the token logits. Generated steps are ensured to not have high similarity with any previous step and to not contradict with the context. The solution path with the highest number of common n-grams with other solutions is selected as the final solution. TreeBoN \citep{qiu2024treebon} utilizes an off-the-shelf DPO model to provide token-level rewards \citep{rafailov2024r} for expanding and pruning a tree of solution paths in best-of-N.

\subsection{Evaluate by External Tools}
Besides using an LLM, external tools can also evaluate the steps. MathDivide \citep{srivastava2024mathdivide} decomposes a problem into a series of sub-problems and converts them into Python programs for evaluating the input values. The outcome of a sub-problem serves as the input to the next sub-problem to obtain the final answer. On the other hand, DTV \citep{zhou2024don} utilizes Isabelle \citep{nipkow2002isabelle} as an automated theorem prover to verify solutions. An informal solution is generated and converted to a formal solution for verification. Multiple solution paths are sampled for verification, and solutions with all reasoning steps verified participate in majority voting to select the final answer. It defaults to Self-consistency \citep{wang2022self} if all solutions failed verification.

\begin{table}
  \centering
  \scriptsize
  \begin{tabular}{|m{2.6cm}|m{2.5em}|m{3cm}|}
    \hline
    \textbf{Dataset} & \textbf{Size} & \textbf{Remarks} \\
    \hline
     AIME & 90 & Competition level\\
     \hline
     AlphaGeometry \newline \citep{trinh2024solving} & 30 & Geometry \\
     \hline
     CollegeMath \newline \citep{tang2024mathscale} & 4K & College-level \\
     \hline
     FrontierMath \newline \citep{glazer2024frontiermath} & 100+ & Expert-level \\
     \hline
     GaoKao2023 \newline \citep{liao2024mario} & 385 & Chinese college entrance exam \\
     \hline
     HumanityLastExam \newline \citep{phan2025humanity} & 1.2K & Expert-level \\
     \hline
     MathOdyssey \newline \citep{fang2024mathodyssey} & 148 & Olympiad-level \\
     \hline
     Math-Shepherd \newline \citep{wang2024math} & 445K & From the MATH and GSM8K datasets; Stepwise labels \\
     \hline
     MathVista \newline \citep{lu2024mathvista} & 661 & College-level; Visual context \\
     \hline
     OlympiadBench \newline \citep{he2024olympiadbench} & 6.1K & Olympiad-level; Bilingual \\
     \hline
     OlympicArena \newline \citep{huang2024olympicarena} & 2K & Olympiad-level \\
     \hline
     Omni-MATH \newline \citep{gao2024omni} & 4.4K & Olympiad-level \\
     \hline
     PRM800K \newline \citep{lightman2023let} & 12K & 800K stepwise labels\\
     \hline
     PRMBench \newline \citep{song2025prmbench} & 6.2K & 83K stepwise labels \\
     \hline
     ProcessBench \newline \citep{zheng2024processbench} & 3.4K & Competition-level \newline Stepwise labels \\
     \hline
     ProofNet \newline \citep{azerbayev2023proofnet} & 371 & College-level Proofs \\
     \hline
     PutnamBench \newline \citep{tsoukalas2024putnambench} & 1.7K & Competition-level Proofs \\
     \hline
     RewardMATH \newline \citep{kim2024evaluating} & 483 & From the MATH dataset \\
     \hline
  \end{tabular}
  \caption{Recent challenging datasets present problems at the college, competition, and Olympiad levels.}
  \label{tab:datasets}
\end{table}

\section{Datasets}
While LLMs perform well on datasets with K-12 difficulty, we present more challenging datasets at the college and competition levels, which tend to be more recent with a lower risk of data contamination. As we summarize them in Table \ref{tab:datasets}, additional details can be found in Appendix \ref{sec::hardproblems}, and K-12 datasets are discussed in Appendix \ref{sec::easyproblems}.

\section{Challenges and Future Directions}
Although multi-step reasoning improves LLMs in solving math problems, it also leads to longer output lengths, which is rather inefficient for easier math problems that can be answered in a short response. To optimize efficiency, an LLM should reason to solve difficult problems while providing direct answers to easier ones, though determining this distinction can be a challenge. Previous datasets typically focus on elementary problems involving basic math operations. As LLMs continue to excel on these benchmarks, it becomes crucial to evaluate them on more complex problems to provide a more comprehensive assessment of their mathematical abilities. Recent datasets address this gap by introducing problems at the college and competition levels (Section \ref{sec::hardproblems}), presenting a more substantial challenge for performance. The same goes for multilingualism. While previous datasets are mostly in English and Chinese, LLMs should generalize across languages. As most papers have yet experimented with complex and multilingual problems, it is important to start incorporating those problems into the benchmark, paving the way to more capable LLMs.

To further improve PRMs and ORMs in multi-step reasoning, it is essential to study the presence of reward hacking \citep{amodei2016concrete}, a phenomenon where the model learns unintended behavior that results in high rewards, such as repeating or reiterating a previous correct step. Such behavior does not make one step closer to the correct final answer, but it can be assigned a high reward due to its correctness to the original question. Another challenge is the inverse scaling law \citep{gao2023scaling, zeng2024scaling} where the policy model learns from feedback from the reward model and deviates away from the original distribution over the training iterations, yet the feedback is provided by the reward model trained on the original distribution, resulting in a distribution shift. This hinders the reward model from providing useful feedback as training prolongs, but the issue can potentially be mitigated by adapting the reward model correspondingly.

Math problems have attracted most of the research efforts in reasoning due to the logical nature of mathematics. The inherent logical property applies to code \citep{ding2024reasoning} and plan \citep{valmeekam2023can} generation problems, but as multi-step reasoning shows promise in these fields, it has the potential to expand to other domains. In fact, the technique has started to appear in many domains, including image captioning \citep{wang2024exovip}, question answering \citep{xue2024enhancing}, and cross-lingual tasks \citep{ranaldi2024tree}. At the highest level, large world models \citep{xiang2024language} can provide multi-faceted feedback from the entire environment to facilitate multi-step reasoning for various tasks beyond mathematical reasoning. 

\section{Conclusion}
This survey presents the various usage of feedback to facilitate multi-step reasoning for LLMs to solve math problems, including step-level and outcome-level feedback. We propose a taxonomy to categorize training-free and training-based approaches and further discuss their subcategories in detail. We present recent datasets with challenging problems to encourage more robust benchmarking, and we discuss other challenges and future directions going forward. In this rapidly evolving field, we aim to establish a solid foundation for readers and make research more accessible and easier to navigate.

\section*{Limitations}
While we make our best efforts to present a comprehensive survey, it is possible that we may have missed some references. Due to the page limit, we choose to describe each work in more relevant parts, so our content can omit further details. For the datasets, we focus on the recent challenging datasets and present other datasets in the Appendix.
\bibliography{main}

\appendix

\section{Survey Methodology}
To systematically obtain relevant references, we first query keywords in Google Scholar, including process rewards, outcome rewards, verifier models, multi-step reasoning, and LLMs. We extend the coverage by adding additional references mentioned in their related work sections. To ensure relevance, we manually examine the methodology of each reference.

\section{Datasets} 

\subsection{K-12 Difficulty} \label{sec::easyproblems}
This section presents math datasets with K-12 difficulty.

AddSub \citep{hosseini2014learning} collects 395 math problems in 1.5K sentences, involving only addition and subtraction at the elementary level.

ASDiv \citep{miao2021diverse} contains 2.3K diverse math problems on basic arithmetic and aggregative operations at the elementary level.

GSM8K \citep{cobbe2021training} includes 8.5K grade-school math problems on basic arithmetic operations, split into 7.5K training and 1K test problems. GSM-hard \citep{gao2023pal} increases the numbers in the questions to create a more difficult version of the test set.

MultiArith \citep{roy2016solving} consists of 600 multi-step arithmetic problems extracted from the commoncoresheets website. The steps are a mixture of addition, subtraction, multiplication, and division.

SVAMP \citep{patel2021nlp} applies variations to existing math problems to create the dataset at the elementary level, resulting in 1K problems with 1.24 operations on average. 


\subsection{College Difficulty and Beyond} \label{sec::hardproblems}
This section presents math datasets at the college and competition levels.

AIME\footnote{\url{https://huggingface.co/datasets/AI-MO/aimo-validation-aime}} presents 90 problems from the American Invitational Mathematics Examination across three years. The exam helps select participants for the international Olympiad.

AlphaGeometry \citep{trinh2024solving} collects 30 Olympiad-level questions on geometry.

AQUA-RAT \citep{ling2017program} involves 100K algebraic problems, each with five choices and a multi-step rationale. The problems are inspired and modified from graduate admission tests.

CollegeMath \citep{tang2024mathscale} contains college-level questions on linear algebra and differential equations, with 1.2K and 2.8K samples in the training and test sets.

FrontierMath \citep{glazer2024frontiermath} consists of hundreds of extremely difficult problems crafted by experts, with the best current model reaching a low accuracy.

GaoKao2023 \citep{liao2024mario} includes 385 challenging problems from the Chinese college entrance exam in 2023, translated into English.

HumanityLastExam \citep{phan2025humanity} presents a challenging benchmark of manually-reviewed questions from various subjects and contains 1.2K expert-level math problems.

MathOdyssey \citep{fang2024mathodyssey} collects 101 and 148 college-level and Olympiad-level problems across 12 domains.

MATH \citep{hendrycks2021measuring} contains 12.5K challenging math problems at the competition level. Each problem has a step-by-step solution, which facilitates model training on step-level annotation.

Math-Shepherd \citep{wang2024math} annotates the MATH \citep{hendrycks2021measuring} and GSM8K \citep{cobbe2021training} datasets to provide 445K problems with explicit step-level annotations. The solutions contain individual steps, and the annotations mark whether each step has the potential to reach the correct final answer.

MathQA \citep{amini2019mathqa} annotates over the AQUA-RAT \citep{ling2017program} dataset by modeling operation programs to provide stepwise solutions, resulting in 37.2K problems.

MathVista \citep{lu2024mathvista} presents a diverse benchmark on questions with visual context, including 661 college-level problems.

MiniF2F \citep{zheng2022minif2f} contains 488 Olympiad-level problems for therorem proving.

OCW \citep{lewkowycz2022solving} consists of 48 undergraduate-level math problems on differential equations from MIT OpenCourseWare.

OlympiadBench \citep{he2024olympiadbench} includes 6.1K Olympiad-level bilingual multimodal problems from math competitions. Each problem includes expert annotations to facilitate step-level reasoning.

OlympicArena \citep{huang2024olympicarena} presents a cognitive reasoning benchmark that includes 2K Olympiad-level math problems.

Omni-MATH \citep{gao2024omni} contains 4.4K Olympiad-level problems from over 33 domains.

PRM800K \citep{lightman2023let} contains 800K manually annotated binary labels on the correctness of each step for generated solutions on the MATH \citep{hendrycks2021measuring} dataset.

PRMBench \citep{song2025prmbench} collects 6.2K problems and 83K step-level labels, measuring the simplicity, soundness, and sensitivity of LLMs.

ProcessBench \citep{zheng2024processbench} consists of 3.4K samples at the competition level, with each containing a human-annotated step-level solution.

ProofNet \citep{azerbayev2023proofnet} includes 371 college-level math proof questions.

PutnamBench \citep{tsoukalas2024putnambench} contains 1.7K formalizations of theorems that are crafted manually from the William Lowell Putnam Mathematical Competition.

RewardMATH \citep{kim2024evaluating} proposes 483 problems from MATH \citep{hendrycks2021measuring} to evaluate the robustness of reward models and enhance upon the math portion of RewardBench \citep{lambert2024rewardbench}, by contrasting a correct solution with nine incorrect ones for each problem.

\end{document}